\documentclass[letterpaper]{article} % DO NOT CHANGE THIS

\usepackage[a4paper, total={6in, 8in}]{geometry}
\usepackage{times}  % DO NOT CHANGE THIS
\usepackage[hyphens]{url}  % DO NOT CHANGE THIS
\usepackage{graphicx} % DO NOT CHANGE THIS
\usepackage[title]{appendix}
\usepackage{algorithm}
\usepackage{algpseudocode}

\usepackage{amsmath}
\usepackage{multirow}
\usepackage{booktabs}
\usepackage{array}
\usepackage{rotating}
\usepackage[table,dvipsnames]{xcolor} 
\usepackage{hhline}
\newcommand{\wtv}{w2v} % Word2Vec
\newcommand{\dtv}{d2v} % Dict2Vec
\newcommand{\ftext}{ftxt} % Fast Text
\newcommand{\gone}{g1} % Glove 100
\newcommand{\gthree}{g3} % Glove 300
\newcommand{\cnnb}{cnnb} % Conceptnet
\newcommand{\wg}{wg} % w2v-glove
\newcommand{\elmo}{elmo} % elmo
\newcommand{\deductive}{$DG_{B}$}
\newcommand{\bayesian}{$BG_{B}$}
\newcommand{\mixed}{$MG_{B}$}
\newcommand{\bayessm}{$ SM_{B} $}

\newcommand{\uncertain}[1]{$\widehat{\mbox{#1}}$}

\newcommand{\cdbar}{\hspace{2pt} | \hspace{2pt}}
\newcommand{\Gk}[1]{G$^{#1}$}
\newcommand{\Sk}[1]{S$^{#1}$}

\newcommand{\gold}{\cellcolor{Goldenrod!70}}

\newcommand{\lightgold}{\cellcolor{Goldenrod!30}}

\title{Improving Cooperation in Language Games with Bayesian Inference and the Cognitive Hierarchy}
\author{
    Joseph Bills, Diego Blaylock, Christopher Archibald \\
}
\date{
    Department of Computer Science, Brigham Young University
    }

\begin{document}

\maketitle

\begin{abstract}
In two-player cooperative games, agents can play together effectively when they have accurate assumptions about how their teammate will behave, but may perform poorly when these assumptions are inaccurate. 
In language games, failure may be due to disagreement in the understanding of either the semantics or pragmatics of an utterance.  
We model coarse uncertainty in semantics using a prior distribution of language models and uncertainty in pragmatics using the cognitive hierarchy, combining the two aspects into a single prior distribution over possible partner types. 
Fine-grained uncertainty in semantics is modeled using noise that is added to the embeddings of words in the language. 
To handle all forms of uncertainty we construct agents that learn the behavior of their partner using Bayesian inference and use this information to maximize the expected value of a heuristic function.  
We test this approach by constructing Bayesian agents for the game of Codenames, and show that they perform better in  experiments where semantics is uncertain.
\end{abstract}

%%%%%%%%%%%%%%%%%%%%%%%%%%%%%%%%%%%%%%%%%%%%%%%%%%%%%%%%%%%%%%%%%%%%%%%%

\section{Introduction}
Many real-world problems require cooperation with AI or human agents for success.
Cooperative games provide an important setting for the development and evaluation of cooperative AI techniques. 
Agents generally utilize some form of communication to cooperate. 
Some settings define the semantics of the communication actions which exist only within the game. 
However, agents can differ in their use of these actions in practice to communicate information. 
One challenge faced by some reinforcement learning approaches to these problems is that agents develop their own way of using actions to communicate during training. 
When these agents interact with other agents who do not use the actions in the same way, performance can suffer greatly. 

In this paper, we focus on a specific type of cooperative game: \emph{cooperative language games}. 
The rules of these games require agents to communicate using natural language. 
One might expect this solves the difficulty agents have communicating, but this is not necessarily the case. 
Because AI agents might use different language models to represent relationships between words, they could use language in the game in different ways. 
Even when agents use the same language model, they may have a different understanding as to what action should be performed in response to an utterance, especially if the meaning of an utterance is ambiguous due to restrictions on the amount of information which can be communicated. 
This illustrates the distinction between the semantic and pragmatic understandings of an utterance, and differences in either understanding can impede communication between agents. 

Our goal is to design an agent for a cooperative language game able to cooperate successfully with any teammate agent, regardless of the semantics of the teammate's language model or the pragmatics of their use of language. 
We approach this using a Bayesian framework, where instead of being restricted to a single language model, our agent is given a set of possible language models over which it can reason probabilistically.
Bayes' rule is used to update the agent's beliefs as it observes its teammate acting. 
This is, to our knowledge, the first example of an agent having a probability distribution over a set of language models, updating this distribution given observations, and adapting in this manner to any individual teammate. 

We explore this idea in the specific cooperative language game Codenames. 
This game, described in Section \ref{sec:codenames}, involves using single words as clues to convey information about the true state of the game. 
The linguistic concepts of \emph{semantics} and \emph{pragmatics} have specific interpretations within the Codenames AI framework. 
In Codenames, semantics corresponds to the language model an agent uses to determine the relationship of words in the game, while pragmatics refers to the strategy an agent uses to determine which action to take, based on their understanding of semantics, their partner, and any signals received.
An example of a difference in semantics would be using word2vec \cite{mikolov2013distributed} instead of GloVe \cite{pennington2014glove} as a word embedding. 
A difference in pragmatics might be guessing as many words as the clue gave in the order of increasing distance versus stopping once there are no more words within a set distance of the clue. 
Most previous Codenames agent models assume a common understanding of pragmatics where a guesser is expected to guess words based on similarity, so their behavior is parameterized only by differences in semantics, but previous work shows alternative pragmatics can improve performance even with teammates using the same language model \cite{bills2023deductive}. 

Our proposed Bayesian framework will model and reason about uncertainty in both semantics and pragmatics within the game of Codenames. 
We will demonstrate that the resulting Bayesian agents have improved performance with a variety of other agents in Codenames. 
The remainder of the paper will provide required background and details of the game of Codenames. 
The Bayesian approach will then be described, followed by the details of our experiments evaluating it.

\section{Background} \label{sec:related}

Much previous research has focused on the task of agents who can cooperate with arbitrary teammates, which has been called \emph{ad hoc teamwork} \cite{mirsky2022survey,stone2010ad}.
The different approaches proposed include some from a reinforcement learning perspective where the goal is to determine a representative set of potential teammates strategies that can be used during training \cite{canaan2019diverse,rahman2024minimum}, while others instead focus on modifying behavior during play with a specific teammate \cite{bard2013online,pedersen2023alegaatr}. 
An example of a game that has been widely used as an ad hoc domain is Hanabi. 
Hanabi is a cooperative card game where players communicate using game-defined actions to gain information about their own cards, so that they can successfully play their cards and benefit the group \cite{bard2020hanabi,canaan2019diverse,canaan2023generating,foerster2019bayesian}. 
This is an example where the semantics of the communication are defined by the rules of the game, and so agents can be assumed to have the same understanding of them. 
The major uncertainty with an unknown teammate comes in the pragmatics of how that agent will use the actions. 
In contrast, Codenames involves natural language used outside the game. 
This means that agents might differ in both their semantic interpretation and representation of words as well as in the pragmatics of how they use the words in the game. 
 
Bayesian statistics provides a framework for reasoning about uncertainty \cite{bolstad2016introduction}, which has been widely applied to multi-agent settings enabling AI agents to reason about teammates and opponents.
As some examples, Bayesian reasoning has been used within the game of Hanabi \cite{foerster2019bayesian,canaan2023generating}, $n$-player competitive games \cite{sturtevant2006prob} and auctions \cite{baarslag2013predicting}, and to determine which of an agent's strategies will work best against the current opponent in poker \cite{bard2013online}.
In each case the set of objects over which probabilistic beliefs are held, the way those beliefs are updated upon observations of events in the game, and how the current beliefs are utilized to determine the best action must be designed specifically for that domain. 
We explore the application of similar ideas to Codenames, which to our knowledge involves the first case of having a probability distribution over a set of language models in a cooperative language game. 

In addition to utilizing Bayesian reasoning to reason about pragmatics we design our proposed agents to fit within a \emph{hierarchy}. 
The model we use is similar to the cognitive hierarchy introduced by Stahl \cite{stahl1993evolution,stahl1995players}. 
Our model differs in that best-responses are only approximate and that in addition to a distribution of lower level agents, the belief model also includes alternate language models.
The cognitive hierarchy itself is an extension of $k$-level reasoning, which has been applied to cooperative games including Hanabi \cite{cui2021klevel}, and the Keynesian beauty contest \cite{shapiro2014level}. 

Semantics and pragmatics are two foundational linguistic concepts.
Semantics refers to the literal meaning of words and phrases and how the meanings of different words are related, while pragmatics refers to how context and the goals of speech acts change the meaning of utterances \cite{grice1957meaning} 
Most computational models of semantics are based on the distribution hypothesis, which states the meaning of words can be inferred from the meaning of other words used in the same context \cite{harris1954distributional}.
Note that while context is used to train these models, many of these models do not model context itself and thus only model semantics. 
A popular form of semantic model is the vector space \emph{word embeddings}, which maps words to points in a normed vector space. 
In addition to capturing the similarity between words through similarity metrics in the vector space, the vector space can also capture semantic relationships between words through vector operations \cite{mikolov2013efficient}. 
All language models we utilize in this work are vector-space word embeddings. 
\cite{lee2021learning} used noise to perturb word embeddings and change words, showing the potential of using random noise to simulate slightly different language models.   
The Rational Speech Act framework \cite{goodman2016pragmatic} models pragmatics as a Bayesian game \cite{harsanyi1967games}, and this is the understanding of pragmatics that guides our model.

\subsection{Codenames} \label{sec:codenames}
Codenames is a board game which explicitly involves language and concepts of similarity between words \cite{codenames}. 
In the game, there are 25 cards with a single word on them laid out on the board. 
Each board card belongs to a hidden category. 
The possible categories are red, blue, bystander, and assassin. 
Two teams (red and blue) of at least two players each, compete to be the first to identify all of their team's board cards.
The players take one of two roles on each team: \emph{spymaster} and \emph{guesser}. 
The spymaster is aware of the hidden category of each word on the board, but only the guesser can pick a board card and reveal its category assignment. 
Each turn, the spymaster gives a clue to the guesser. 
Each clue consists of a single word and a number. 
The clue word is related in meaning to some of the cards on the board, while the clue number is generally the number of board cards the spymaster intends to connect to the clue word.
The guesser must guess at least one card, revealing its hidden category, and picks cards until either 1) a card is guessed that doesn't belong to their team, 2) they have guessed one card more than the clue number, or 3) they choose to end their turn.
A team wins when all of their cards have been revealed, ending the game. A team loses instantly if they guess the assassin card.

A single-team variant of Codenames, where the goal is to guess all the team's cards in as few turns as possible, was used in a Codenames AI competition, and has been used in subsequent AI research. on Codenames \cite{competition, kim2019cooperation}. 
The best AI Codenames agents to date have used \emph{word embeddings}. 
While alternatives have been explored, such as large language models \cite{costarelli2024gamebench}, they have not yet been as effective as the embedding approaches.
Embedding approaches have a modular design, making it easy to separate and interchange the semantic and pragmatic elements of an agent.
An agent's word embedding defines the relationships between words, and basic guessers essentially guess board cards in the order of distance from the clue word, according to their word embedding. 
Spymasters give the clue that maximizes the number of board cards correctly identified by a basic guesser using the same word embedding. 
These basic strategies result in agents that play very well together when using the same word embedding, but performance is typically very poor when embeddings differ \cite{kim2019cooperation}. 
Subsequent work generally expanded the explored set of language models and began evaluating with a small set of humans \cite{jaramillo2020word}.
One agent explored in that work used a n\"{a}ive Bayes filter to classify words, but it did not use Bayesian inference to adapt to the behavior of its teammate. 
Other focused on improving clues for play with humans, but didn't use the actual game of Codenames \cite{koyyalagunta2021playing}. 
All the previous work focuses on identifying the single language model and/or strategy that will work best with a given population of teammate agents, oftentimes humans. 
However, all of the proposed agents are static, as they do not adjust or adapt their behavior based on the interaction with their current teammate. 
The baseline agents used in our experiments are derived from these works. 

Conceptually, these agents have an internal model of their partner, and to play well this model needs to be accurate. 
Another paper explored new strategies or pragmatics for an abstraction of Codenames, using a deductive hierarchy where pairs of agents higher in the hierarchy perform better than those lower in the hierarchy. 
On the other hand, agents who are not adjacent in the hierarchy perform poorly with each other due to their inaccurate beliefs about the other \cite{bills2023deductive}. 
A major remaining challenge from all prior Codenames AI research is the creation of agents that can play well despite uncertainty about their partner. 

\section{Foundations} \label{sec:preliminaries}
In this section we introduce the foundational concepts and a general overview of our proposed Bayesian approach for cooperative language agents. 
In particular we will detail how uncertainty regarding both semantics and pragmatics will be represented within the framework. 
Codenames is used throughout as a concrete example of a cooperative language game, but the same concepts should be applicable to similar settings, albeit with some necessary adaptation. 

\subsection{The Deductive Hierarchy} \label{sec:hierarchy}
The deductive hierarchy is an organization of agents for playing Codenames initially proposed in \cite{bills2023deductive}. 
The foundation of the hierarchy (level 0) is a \emph{static guesser}, or a guesser that only considers the current clue and unrevealed board cards in determining its guess. 
The hierarchy then consists of alternating spymasters and guessers, designated so that a level $k$ spymaster (\Sk{k}) approximates a best response to a level $k$ guesser (\Gk{k}), while a level $k$+1 guesser (\Gk{k+1}) similarly approximates a best response to \Sk{k}.
Roughly speaking, \Sk{k} assumes it is playing with \Gk{k}, simulating \Gk{k}'s response to any clue it could give, and choosing the clue that results in the maximum revealed team cards.
\Gk{k} assumes that the clues it receives are generated by \Sk{k-1}. 
\Gk{k} maintains beliefs over all possible states of the world (board card assignments), removing those that are inconsistent with given clues and revealed information. 
Board card identities are deduced when they are true in every remaining possible state.
When \Gk{k} has deduced that a board card is not on its team, it will skip over that card when guessing, and when it deduces a card is on its team it will use the extra guess to guess it. 
This behavior allows hierarchical agents on the same level to gain the most information possible from each clue and win the game in fewer turns. 
From the perspective of this hierarchy, the basic Codenames AI agent framework initially described in \cite{kim2019cooperation} consists of a level 0 guesser and a level 0 spymaster. 
Agents from higher levels in the hierarchy are dynamic, as the clues and guesses they produce will depend upon the entire history of the game to that point. 
The deductive hierarchy is fragile and hierarchical agents perform poorly when assumptions about teammates are incorrect. 
Our proposed Bayesian framework can be viewed as an extension of the hierarchy to reason probabilistically and be more robust with all teammates. 

\subsection{Modeling Uncertainty: Pragmatics and Semantics}
One of the core ideas of any Bayesian framework is to explicitly model and account for sources of uncertainty. 
The main sources of uncertainty in cooperative language games like Codenames are semantics and pragmatics. 
We represent uncertain semantics by a probability distribution over a set of different word embeddings.
Uncertain pragmatics can be represented by a probability distribution over different levels of the just described deductive hierarchy. 
Both of these sources of uncertainty can be captured by having a set of possible teammates, each with a word embedding and level in the deductive hierarchy. 
In order to account for the possibility of partnering with an unknown agent, we also add a noise model to the word embeddings.  
This means that any clue or guess has some non-zero probability of being generated by any agent model. Based on prior work involving perturbed word embeddings \cite{lee2021learning}, we use Gaussian noise to simulate uncertainty in the implied communication channel.  

\subsection{Bayesian Approach Overview}\label{sec:overview}
We now provide an overview of the proposed Bayesian approach for Codenames agents.
Each agent will have a set of possible teammate models $M$, and each $m \in M$ should differ by the word embedding it is using (semantics) or its position in the hierarchy (pragmatics). 
Beliefs over these teammate models will be maintained in the form of a probability distribution $P(m)$.
When a teammate action $a$ is observed, the agent will update its beliefs using Bayes rule as $P(m \cdbar a) \propto P(a \cdbar m)P(m)$, where the posterior $P(m \cdbar a)$ will be used as the prior $P(m)$ for the next update.

The distribution $P(a \cdbar m)$ corresponds to the specifics of how teammate model $m$ acts in the game, as determined by its strategy and word embedding. 
The Bayesian approach will fail if $P(a \cdbar m) = 0$ for all teammate models when action $a$ is observed since it will result in the posterior $P(m)$ being set to zero for all models, preventing any future inference.   
% This can occur when playing with a teammate that is not exactly represented in the set $M$.
To avoid this problem and ensure all models may continue to be used for inference despite faults in the approximation, each teammate model should have a non-zero probability of generating any action.
All previous Codenames agents of which we are aware have been deterministic: for the same state of the game, they would generate the same guess or clue. 
Thus, these previous agent designs must be modified or adapted to be stochastic before they can be used effectively as teammate models in our proposed Bayesian approach.
This can be done by adding a random perturbation the embeddings of words, and details of this process will be provided in later sections. 

\subsection{Heuristic Utility Function} \label{sec:utility}
When making a decision in the game, Bayesian agents will use their beliefs to select the action that maximizes expected utility. 
The Bayesian agents will utilize the following heuristic utility function which will provide a utility for a sequence of cards to be guessed on one turn, based on the identity of revealed cards. 
The spymaster, with knowledge of the true world state, can use the actual card identities, while the guesser will instead use a possible world state.
The utility of a turn is the sum of the values of any cards revealed that turn, minus 1.
% $U(g) = \sum_{c \in g} u(c) - 1$
The value for the card types are as follows, assuming the agents are on the red team:  
$u(\mbox{red}) = 1$,
$u(\mbox{blue}) = -1$,
$u(\mbox{bystander}) = 0 $,
$u(\mbox{assassin}) = -|R|$, 
where $|R|$ is the total number of red team cards.
This heuristic calculates the marginal contribution to the score at the end of the game from the current turn assuming a particular variant of the solitaire Codenames where an opponent card must be revealed each turn. 
Further explanation is found in the appendix. 
The heuristic utility function could easily be replaced by another, where motivated. 

We now describe the Bayesian spymaster, followed by the Bayesian guesser. 
Due to space restrictions, the full details of each agent will not be given here, but details are included in the appendix and source code. 
The experimental evaluation of the agents will be given in Section \ref{sec:experiments}. 

\section{The Bayesian Spymaster} \label{sec:spymaster}
The Bayesian spymaster aims to deduce which guesser it is playing with, so that it can give more effective clues.
The Bayesian spymaster maintains beliefs over a set of guesser models $M$, represented by a probability distribution that is updated after each turn and thus incorporates all information obtained by previous guesses.
Initially, these beliefs are set to be the uniform distribution over $M$. 
Given observation of a guess $g$, the update is $P(m \cdbar g) \propto P(g \cdbar m) P(m)$. 
The conditional probability $P(g \cdbar m)$ cannot be computed exactly for arbitrary guessers, so it is estimated with a multinomial distribution, using Monte Carlo sampling to count guess occurrences for each model and Laplace smoothing initializing all the counts to 1.  
Each guesser model is sampled multiple times, where each time noise is added to the guesser's embedding of the clue word. 
Distances to board words are computed from the perturbed embedding and the guesser then guesses according to its strategy. 
Since all guesses are included in the support for all guessers, a Bayesian spymaster is never completely certain which guesser it is paired with. 

To determine which clue it should give, the Bayesian spymaster computes the expected utility of each potential clue with each model guesser. 
A similar sampling process is used where each model guesser is queried multiple times, and its response generated using the perturbed embedding of the clue word. 
The total expected utility of a clue is then calculated for each guesser across these samples and across all models using the current beliefs. 

Clue words are omitted from consideration if, without noise, each guesser model in $M$ would guess an incorrect card, given that clue. 
While this optimization sacrifices theoretical optimality, it greatly increases the computational speed of the agent and performs well in practice. Further details are in the appendix.

If the Bayesian spymaster uses no noise and has only a single level-$k$ guesser model, then it is equivalent to a level-$k$ spymaster from \cite{bills2023deductive}. 

\section{The Bayesian Guesser} \label{sec:guesser}
The Bayesian guesser will have a set of spymaster models, $M$, over which it will maintain beliefs. 
It also maintains a history of previous clues and is parameterized with thresholds for guessing and skipping cards. 
After receiving a clue, the guesser will first sample possible world states consistent with the current observed state of the board using the method described in \cite{bills2023deductive}. 
For each clue $l_t$ in the history, it will calculate the likelihood the clue was given for each combination of spymaster $m$ and world state $w$ in the sample. 
This likelihood is given exactly by $\int_{\nu(l)}\mathcal{N}(m_t(w),\sigma,x)dx$ where $\nu(l)$ is the Voronoi region centered around $l$, $m_t(w)$ is the clue model $m \in M$ would give at turn $t$ assuming $w$ was the true state of the world, and $\mathcal{N}(m_t(w),\sigma,x)$ is the multivariate symmetric Gaussian centered at $m_t(w)$ with $\sigma$ covariance where $x$ is an arbitrary point in the embedding space.  
The Gaussian distribution is used here since it is the noise added to the embeddings in our experiments, but it could be replaced by another distribution. 
The entire history's likelihood is calculated as the product of the likelihoods for each turn's clue. 
The likelihoods of the clue history for each model and world state can be separately marginalized out. 
If there is a nonzero likelihood of the current clue occurring then the beliefs about model probabilities are updated according to the posterior distribution. Otherwise this step is skipped, mirroring the behavior of the level-$k$ guesser \cite{bills2023deductive}. 
If beliefs are not updated then the clue is also not added to the history. 

Next, the posterior beliefs of each world state are calculated assuming they were equally likely a priori. 
The probability of each unknown card on the board belonging to any category is then estimated using a n\"{a}ive Bayesian filter. 
Board cards are first filtered using the \emph{skip threshold}.
Any card with a probability of being on the guesser's team equal to or below the skip threshold is removed from consideration for boosting it's probability beyond what a strict Bayesian interpretation would infer. 
Among the remaining board cards, the $n$ closest to the clue have their probability of being their team’s card \emph{boosted}, or increased, to one, where $n$ is the numerical component of the clue and distance is measured according to the word embedding of the model spymaster with the highest posterior probability.  

Board cards are then filtered using the \emph{belief threshold}. Among the cards where the probability of being on the team is at least equal to the belief threshold, the card with the highest positive expected utility according to the heuristic function of Section \ref{sec:utility} is selected. 
If no cards have a utility greater than zero, then the card with the highest expected utility among all cards is selected instead, since the rules require at least one card to be guessed.

To complete the remainder of the guess, the guesser assumes the first guessed card belongs to the guesser's team, then repeats the process of filtering cards with the value of $n$ deceased by one.  This process stops when all possible guesses have been used or when there are no cards with a positive expected heuristic value and a probability of being on the team greater than the belief threshold, in which case the guesser ends its turn. 

Finally, if the guesser ever observes a card assignment which it believed had a zero probability of occurring then it will reset, again mirroring the behavior of the level-$k$ guesser. 

\subsection{Skip and Belief Thresholds}
To ensure the Bayesian Guesser is a strict generalization of the level-$k$ guesser, it is parameterized using aforementioned skip and belief thresholds. 
In the deductive hierarchy, proximity to the clue is a signal used to signify which cards are on a player's team, but clues which have been deduced to not be on the player's team are skipped when updating beliefs using this signal. 
This behavior is not directly modeled in the Bayesian ideal and is missing in some edges cases, so it is explicitly modeled in the Bayesian Guesser by having the probability of close cards be boosted. 
As such, the skip threshold describes the posterior probability where close cards are considered to not have be signalled because they are likely not on the player's team.
The belief threshold describes how confident the guesser must be that the card is on the team before it is guessed. 
These parameters are designed so that when belief = 0 and skip = 1  the model behaves uses the ideal Bayesian reasoning described, but as skip $\xrightarrow{}$ 0 and belief $\xrightarrow{}$ 1 it relies less on the heuristic and Bayesian reasoning and more on an approximation of logical deduction. 
When skip = 0 and belief = 1, there is no noise, and the set of modeled guessers is a singleton of the level-$k-1$ guesser, it behaves identically to the level-$k$ guesser.
Details for how these parameters are integrated into the model are included in the appendix.

\subsection{Cognitive Hierarchy}
A cognitive hierarchy can be constructed by assuming Bayesian agents have prior distributions over other Bayesian agents. 
To define this behavior, whenever a reference is made to the word embedding of a Bayesian agent, the word embedding of its highest posterior probability agent becomes the assumed word embedding, with a defined ordering of agents to break ties. 
This implies the word embedding of a level $k$ agent is the same as the level $0$ guesser it is founded on. 
This allows arbitrary cognitive hierarchy models to be defined using Bayesian agents. 

\section{Experimental Evaluation in Codenames} \label{sec:experiments}
We now describe the details of the Codenames experiments that were carried out to evaluate the Bayesian Codenames agents just described at level $k=1$ in the hierarchy.

\subsection{Varied Semantics: Word Embeddings}
The word embeddings used in the experiments include: 
    \begin{itemize}
        \item Word2Vec (\wtv{}) -- trained using a word context windows \cite{mikolov2013distributed}.
        \item Dict2Vec (\dtv{}) -- similar to \wtv{} but trained on cleaned dictionary entries with an improvement on semantic similarity tasks \cite{tissier2017dict2vec}.
        \item FastText (\ftext{}) -- Uses bags of character $n$-grams with weighting by position \cite{mikolov2018advances}.
        \item GloVe (\gone{}, \gthree{}) -- trained on pre-computed statistical co-occurrence probabilities for words in a corpus \cite{pennington2014glove}. We used embeddings with dimensions 100 and 300.
        \item ConceptNet Numberbatch (\cnnb{}) -- uses retrofitting to incorporate the ConceptNet Knowledge graph into an embedding. \cite{speer2017conceptnet}.
        \item Word2Vec-Glove (\wg{}) -- concatenation of \wtv{} and 50-dimensional GloVe that previous work found effective \cite{kim2019cooperation}.
        \item ELMo (\elmo{}) -- a 1024-dimensional de-contextualized embedding derived from 3 layers of a trained contextual model \cite{peters2018deep}. 
        The context-free embedding was created by pooling contextual embeddings across many contexts \cite{bommasani2020interpreting}.
    \end{itemize}

\subsection{Experiment Setup}
For the Bayesian Spymaster, we used two models referred to as \bayessm{} and \uncertain{\bayessm{}} the first assuming zero perturbation noise, and the second assuming a noise value of 1.0. 

6 different Bayesian guessers were evaluated: \deductive{}, \uncertain{\deductive{}}, \bayesian{}, \uncertain{\bayesian{}}, \mixed{}, and \uncertain{\mixed{}}. 
Each of the three types again had a version without embedding perturbation noise and a version with a noise value of 1.0. 
The three types differ in their skip and belief threshold parameters with \deductive{} being set to skip = 0 and guess = 1, \bayesian{} set to skip = 1 and guess = 0, and finally \mixed{} set to skip = 0.5 and guess = 0.5.

All the Bayesian agents included the same set of word embeddings (\wtv{}, \gthree{}, \cnnb{}, and \dtv{}) in the model set $M$.
These are called the \emph{internal} word embeddings. 
A set of static non-Bayesian agents using these word embeddings was included in each experiment as a baseline.  
In addition, the following word embeddings were also used in the experiments, although they were not part of the Bayesian agents: \gone, \ftext, \wg, and \elmo. 
This set of word embeddings will be referred to \emph{external}. 

To more efficiently calculate the set of possible clues, the 300 nearest neighbours of each word were precomputed. 
The probability that a perturbed vector would fall in the Voronoi region for any clue was precomputed using 1000 samples at each noise levels. 
This was done by perturbing a model's word embedding using a standard normal distribution with mean at the embedding and then finding the closest word among the 500 closest neighbours. 
The Bayesian spymasters all used 10 samples, and the Bayesian guessers used 1000 or 10,000 samples.

Each of the Bayesian spymasters -- as well as the static spymasters for both internal and external word embeddings -- were evaluated in environments both with and without the addition of stochastic embedding perturbations to all communicated clue words.  
These are called \emph{stochastic} and \emph{deterministic} environments respectively. 
Each spymaster played against all of the guessers. 
The guessers consisted of all the Bayesian guessers, as well as static guessers using both internal and external word embeddings, in both stochastic and deterministic environments. 
In stochastic environments noise was added to the clue embedding for the guesser in the Bayesian spymaster experiments, 
while for the Bayesian guesser experiments noise was added to the spymaster's word embedding and then transformed to the closest clue in the vocabulary before being passed to the guesser.  
Each pairing played 500 games.
The results report the \emph{win rate} for each pair, which is the fraction of solitaire games the pairing is able to successfully win.

\renewcommand{\arraystretch}{1.4}
\setlength{\tabcolsep}{6pt}
\begin{table*}[t]\centering
\caption{Experimental Win Rate Results for Bayesian Spymasters}\label{tab:b-spymasters}
\begin{tabular}{|c|c|cccc|c|cccc|c|c|}
\cline{3-12}
\multicolumn{2}{c|}{} &\multicolumn{5}{c|}{\textbf{In-distribution guessers}} &\multicolumn{5}{c|}{\textbf{Out-of distribution guessers}} \\\hhline{~~----------~}
\multicolumn{2}{c|}{} &\textbf{w2v} &\textbf{g3} &\textbf{cn} &\textbf{d2v} &\textbf{Avg} &\textbf{g1} &\textbf{ftxt} &\textbf{wg} &\textbf{elmo} &\textbf{Avg} \\ \hline
\multirow{3}{*}{ \begin{sideways} \textbf{Det. Env.} \end{sideways}} &Best Model &1.000 \lightgold{}&1.000 \lightgold{}&1.000\lightgold{} &0.702 &\textbf{0.926} &0.828 &0.61 &0.894 &0.538 &\textbf{0.718} \\  %\hhline{~-----------~}
&\textbf{\bayessm{}} &1.000 \lightgold{} &1.000 \lightgold{}&1.000 \lightgold{}&0.858\lightgold{} &\textbf{0.960} \gold{} &0.840 &0.656 &0.956 &0.592 &\textbf{0.761} \\ %\hhline{~-----------~}
&\textbf{\uncertain{\bayessm{}}} &0.830 &0.994 &0.954 &0.650 &\textbf{0.860} &0.974 \lightgold{}&0.802 \lightgold{}&0.990 \lightgold{}&0.720 \lightgold{}&\textbf{0.872} \gold{}\\ 
\hline
\hline
\multirow{3}{*}{ \begin{sideways} \textbf{Stoch. Env.} \end{sideways}} &Best Model &0.364 &0.568 &0.184 \lightgold{}&0.364 &\textbf{0.370} &0.592 &0.208 &0.606 &0.464 &\textbf{0.468} \\ %\hhline{~-----------~}
&\textbf{\bayessm{}} &0.398 &0.614 &0.140 &0.324 &\textbf{0.370} &0.568 &0.224 &0.650 &0.464 &\textbf{0.477} \\ %\hhline{~-----------~}
&\textbf{\uncertain{\bayessm{}}} &0.438 \lightgold{}&0.846 \lightgold{}&0.140 &0.380 \lightgold{}&\textbf{0.450} \gold{}&0.810 \lightgold{}&0.232 \lightgold{}&0.850 \lightgold{}&0.608\lightgold{} &\textbf{0.625} \gold{}\\
\hline
\end{tabular}
\end{table*}
\renewcommand{\arraystretch}{1.0}
\setlength{\tabcolsep}{6pt}

\subsection{Experimental Results}

Table \ref{tab:b-spymasters} shows the win-rate performance of the two Bayesian spymasters against different groups of guessers. 
In each case, we compare to the performance of the best static spymaster, using one of the internal word embeddings. 
The left half of Table \ref{tab:b-spymasters} shows performance against guessers that are \emph{in-distribution}, meaning that each guesser is using an internal language model, or one in the spymaster's set of models. 
The right half of the table shows the performance when the guessers are out-of-distribution -- meaning they used external models that the Bayesian spymaster does not have. 

\renewcommand{\arraystretch}{1.4}
\setlength{\tabcolsep}{5.7pt}
\begin{table*}[htp!]\centering
\caption{Experimental Win Rate Results for Bayesian Guessers}\label{tab:b-guessers}
\begin{tabular}{|c|c|cccc|c|cccc|c|c|}
\cline{3-12}
\multicolumn{2}{c|}{} &\multicolumn{5}{c|}{\textbf{In-distribution spymasters}} &\multicolumn{5}{c|}{\textbf{Out-of distribution spymasters}} \\ \hhline{~~----------~}
\multicolumn{2}{c|}{} &\textbf{w2v} &\textbf{g3} &\textbf{cn} &\textbf{d2v} &\textbf{Avg} &\textbf{g1} &\textbf{ftxt} &\textbf{wg} &\textbf{elmo} &\textbf{Avg} \\ \hline
\multirow{7}{*}{ \begin{sideways} \small \textbf{Deterministic Environment} \end{sideways}} &Best Model &1.000 \lightgold{}&1.000 \lightgold{} &1.000 \lightgold{}&0.702 &\textbf{0.926} &0.752 \lightgold{}&0.788 \lightgold{} &0.702 \lightgold{}&0.580 \lightgold{}&\textbf{0.706} \gold{}\\  %\hhline{~~----------~}
&\textbf{\deductive{}} &0.994 &0.990 &0.954 &0.960 \lightgold{} &\textbf{0.975}  \gold{} &0.447 &0.644 &0.564 &0.479 &\textbf{0.534} \\ %\hhline{~~----------~}
&\textbf{\uncertain{\deductive{}}} &0.984 &0.994 &0.940 &0.928 &\textbf{0.962} &0.458 &0.609 &0.542 &0.479 &\textbf{0.522} \\ %\hhline{~~----------~}
&\textbf{\bayesian{}} &0.788  &0.794 &0.794 &0.756 &\textbf{0.783}  &0.356 &0.471 &0.432 &0.374 &\textbf{0.408} \\ %\hhline{~~----------~}
&\textbf{\uncertain{\bayesian{}}} &0.792 &0.786 &0.826 &0.758 &\textbf{0.791} &0.423 &0.518 &0.419 &0.352 &\textbf{0.428} \\ %\hhline{~~----------~}
&\textbf{\mixed{}} &0.904  &0.884 &0.854 &0.878 &\textbf{0.880}  &0.425 &0.607 &0.538 &0.484 &\textbf{0.514} \\ %\hhline{~~----------~}
&\textbf{\uncertain{\mixed{}}} &0.898 &0.910 &0.824 &0.868 &\textbf{0.875} &0.458 &0.587 &0.479 &0.451 &\textbf{0.494} \\ 
\hline
\hline
\multirow{7}{*}{ \begin{sideways} \small \textbf{Stochastic Environment} \end{sideways}} &Best Model &0.720 &0.990 \lightgold{}&0.482 &0.640 &\textbf{0.708} &0.712\lightgold{} &0.450 \lightgold{}&0.688 \lightgold{}&0.580 \lightgold{}&\textbf{0.608} \gold{} \\ %\hhline{~~----------~}
&\textbf{\deductive{}} &0.712  &0.972 &0.262 &0.734 &\textbf{0.670}  &0.397 &0.300 &0.514 &0.479 &\textbf{0.423} \\ %\hhline{~~----------~}
&\textbf{\uncertain{\deductive{}}} &0.704 &0.976 &0.508 &0.742 &\textbf{0.733} &0.417 &0.283 &0.521 &0.477 &\textbf{0.425} \\ 
 %\hhline{~~----------~}
&\textbf{\bayesian{}} &604  &0.772 &0.216 &0.602 &\textbf{0.549}  &0.313 &0.227 &0.419 &0.374 &\textbf{0.333} \\ %\hhline{~~----------~}
&\textbf{\uncertain{\bayesian{}}} &0.638 &0.776 &0.530 &0.648 &\textbf{0.648} &0.363 &0.222 &0.389 &0.348 &\textbf{0.331} \\ %\hhline{~~----------~}
&\textbf{\mixed{}} &0.696  &0.872 &0.252 &0.682 &\textbf{0.626}  &0.389 &0.296 &0.499 &0.484 &\textbf{0.417} \\ %\hhline{~~----------~}
&\textbf{\uncertain{\mixed{}}} &0.770 \lightgold{}&0.894 &0.556 \lightgold{}&0.738 &\textbf{0.740} \gold{} &0.406 &0.274 &0.497 &0.451 &\textbf{0.407} \\ 
\hline
\end{tabular}
\end{table*}
\renewcommand{\arraystretch}{1.0}
\setlength{\tabcolsep}{6pt}

Table \ref{tab:b-guessers} shows the same type of results, but for the Bayesian guessers across different spymasters.

\section{Discussion}

In general, the Bayesian spymaster that assumed noise performed better than any other spymaster whenever they were partnered with out-of-distribution models or in stochastic environments. 
We believe this is because the spymaster rapidly learns which of its model guessers is the best fit for the guesser it is playing with, and then performs better than the corresponding spymaster for that word embedding because it hedges its clues around uncertainty from noise or unmmodeled behavior. 

Surprisingly, the Bayesian spymaster performed better with the guesser using d2v than the corresponding spymaster even in the deterministic environment, which is the condition the baseline spymaster was designed to be approximately optimal in. 
We believe that this is because the Bayesian spymaster can discriminate between different cards types not on the team, allowing it to avoid the assassin in cases where it is forced to give a bad clue, while the baseline spymaster cannot discriminate in this manner.
The average column summarizes these results, showing the overall performance of each spymaster across all guessers.
For both in-distribution and out-of-distribution guessers, in both deterministic and stochastic environments, the Bayesian spymasters have the best average performance.  
With out-of-distribution guessers, \uncertain{\bayessm{}}, the Bayesian spymaster with noise, performs the best by a wide margin. 
The improvement over the best model in these cases represents a huge improvement over previously published Codenames agents in the same cross-language model setting.  
In particular, it is noteworthy that the Bayesian framework allows the spymaster to do better than \emph{any} of its constituant models would do on its own.

In contrast to the Bayesian spymasters, the Bayesian guessers almost never performed as well as the best model. 
We believe this is because the guesser inherently learns slower than the spymaster due to having a weaker signal. Because the clue signals both information about the state of the board and about the identity of the spymaster to the guesser, there is less information to specify the particular spymaster. 

Despite their weaknesses, the Bayesian guessers performed decently in many cases, and for both in-distribution cases one of the Bayesian guessers had the best average performance.
This confirms that while they are not optimal, they are still robust as they were designed to be. 
\deductive{}, the guesser that did not assume noise and had the highest belief and lowest skip threshold did the best of all models on average in the deterministic environment.
We believe this is because this model learns the fastest due to assuming a stronger signal from the lack of modeled noise, and from the fact its deduction behavior discretely changes once a model raises to having the highest posterior distribution. 
Even though these deductions may be less accurate than the other models, the speed with which it acted proved more important in this particular game. 

\section{Conclusions} \label{sec:conclusion}

Bayesian inference and cognitive hierarchies can improve cooperation in language games.
We demonstrated the first use of Bayesian reasoning to adapt to teammates in Codenames.
The Bayesian spymaster was shown to be  especially successful with out-of-distribution teammates, which has been a key difficulty with previous Codenames AI approaches.
Effective implementation requires game-specific optimizations.
While the theoretical description allows for arbitrarily large cognitive hierarchies, practical difficulties in implementing higher levels in the hierarchy made experimentation beyond the first level out of scope for this study. 
We hope to overcome that limitation in the future.
Additionally, all experiments to date have involved only simulated language models. We would like to see how agents perform with human subjects, who display both complex semantic and pragmatic reasoning. We hope work with multiple semantic and pragmatic models in simulated environments will lead to creating agents that can better communicate with people.  

\bibliography{refs}
\bibliographystyle{abbrv}

\begin{appendices}

\section{The Bayesian Guesser} \label{app:guesser}
In this section we give the complete algorithm for the Bayesian Guesser
First, Algorithm \ref{alg:bg-init} is called at beginning of game to initialize the Bayesian guesser.

\begin{algorithm}[H]
\caption{Initialize Bayesian Guesser} 
\label{alg:bg-init}
\begin{algorithmic}
    \Require Set of spymasters $M$, prior probability distribution over spymasters $q(M)$, posterior distribution of guesses $p(M)$, clue history $l$
    \State $l \gets empty$
    \For{$m \in  M$}
        \State $p(m) \gets q(m)$
    \EndFor
\end{algorithmic}
\end{algorithm}

When a clue is received by the Bayesian guesser, Algorithm \ref{alg:bg-run} is then called.
Several sub-algorithms are utilized in the operation of Algorithm \ref{alg:bg-run}. 
These sub-algorithms are shown in Algorithms \ref{alg:world-state}, \ref{alg:voronoi}, \ref{alg:update-model}, \ref{alg:compute-guess}, \ref{alg:first-guess}, and \ref{alg:card-prob}. Algorithm \ref{alg:update-bg} is called after each guessed card is revealed, incrementing $i$, in order to update the Bayesian guesser.

\begin{algorithm}[H]
\caption{Run Bayesian Guesser}
\label{alg:bg-run}
\begin{algorithmic}
    \Require Set of spymasters $M$, posterior probability distribution over spymasters $p(M)$, clue history $l_{1:t}$
    \State $W \gets$ up to $n$ sampled consistent world states
    \State $p(w|l) gets \mathtt{ComputeWorldStateProbabilities}(W,l_{1:t-1},M)$
    \State $\mathtt{UpdateModelProbabilities}(W,l_t,M)$
    \State $g \gets \mathtt{ComputeGuess}(W,M)$
    \State $o = \mathtt{Result}(g)$
    \Comment Execute guess in the real game
    \State $do some things$
\end{algorithmic}
\end{algorithm}

\begin{algorithm}[H]
\caption{Compute World State Probabilities }
\label{alg:world-state}
\begin{algorithmic}
    \Require Set of spymasters $M$, probability distribution over spymasters, clue history $l_{1:t-1}$, sample of world states $W$
    \For{$w \in W$}
        \State{$p(w|l) \gets 1$ }
    \EndFor
    \For{i < t}
        \If{$l_i \neq null$}
            \For{$w \in W$}
                \State{$p_t(w|l_i) \gets 0$} 
                \For{$m \in M$} 
                    \State{$l*_i \gets m_i(w)$}
                    \State{$p(l|m,w) \gets \mathtt{VoronoiProbability}(l_i,l*_i, m_i)$}
                    \State{$p_t(w|l_i) \gets p_i(w|l_i) + p(l|m,w)$}
                    
                \EndFor
                \State{$p(w|l) * p_i(w|l_i)$}
            \EndFor
        \EndIf{}
    \EndFor{}
    \State{\Return {$p(l|w$)}}
\end{algorithmic}
\end{algorithm}

\begin{algorithm}[H]
\caption{Voronoi Probability}
\label{alg:voronoi}
\begin{algorithmic}
    \Require Spymaster $m$, observed clue $l$, intended clue $l*$, number of samples n, noise $\sigma$, vocabulary $V$
    \State{$count \gets 0$}
    \For{$ i < n$}
        \State{$y \sim G(m.model(l*.word), \sigma)$}
        \State{$ d* \gets \infty$}
        \State{$closest \gets null$}
        \For{$v$ in $V$}
            \State{$d \gets \lvert{m.model(v) - y}\rvert$}
            \If{$d < d*$}
                \State{$d* \gets$ d}
                \State{$closest \gets v$}
            \EndIf{}
        \EndFor
        \If{$closest = l$}
            \State{$count \gets count + 1$}
        \EndIf
    \EndFor
    \State{\Return{$count/n$}}
\end{algorithmic}
\end{algorithm}

\begin{algorithm}[H]
\caption{Update Model Probabilities}
\label{alg:update-model}
\begin{algorithmic}
    \Require Set of spymasters $M$, probability distribution over spymasters, clue history $l$, current clue $l_t$, sample of world states $W$
    \State{p = 0}
    \For{$m \in M$}
        \State{$p_t(m|l_t) \gets 0$} 
    \EndFor
    \For{$w$ in $W$}
        \State{$p_t(w|l_t) \gets 0$} 
        \For{$m$ in $M$} 
            \State{$l*_t \gets m_t(w)$}
            \State{$p(l|m,w) \gets \mathtt{VoronoiProbability}(l_t,l*_t, m_t)$}
            \State{$p_t(w|l_t) \gets p_t(w|l_t) + p(l|m,w)$}
             \State{$p_t(m|l_t) \gets p_t(m|l_t) + p(l|m,w)$}         \State{$p \gets p + p(l|m,w)$}  
        \EndFor
    \EndFor{}
    \If{$p = 0$}
        \State{$l.append(null)$}
        \State{\Return {$p(l|w$)}}
    \Else{}
         \State{$l.append(l_t)$}
         \State{$p(m) \gets p(m) * p_t(m|l_t)$} \Comment{Update beliefs}
        \State{\Return {$p(l|w) * p_t(w|l_t)$}}
    \EndIf
      
\end{algorithmic}
\end{algorithm}

\begin{algorithm}[H] 
\caption{Compute Guess}
\label{alg:compute-guess}
\begin{algorithmic}
    \Require Sample of world states $W$, distribution over world states $p(w)$, leading model $m$, clue $lt$, unknown cards on board $B$, guess threshold $u$, guessed card probabilities $P$
    
        \State{$order \gets []$ (empty list)}
    \For{$c \in B$}
        \If{$order.isEmpty$}
            \State{$order.add(c)$}
       \ElsIf{$\lvert m.model(order[order.size -1]) - 
 l_t.vecor\rvert \leq \lvert m.model(c) - l_t.vecor\rvert $}
        \Else
             \For{i < order.size}
                \If{$\lvert m.model(order[i]) - 
 l_t.vecor\rvert \leq \lvert m.model(c) - l_t.vecor\rvert $} 
                \State{order.insert(c,i}
                    \State{break}
                \EndIf{}
            \EndFor{}
        \EndIf
    \EndFor
    \State{$k \gets l_t.number$}
    \State{$p \gets \mathtt{Card Probabilities}(W,order.copy,k)$}
    \State{$c* \gets FirstGuess(p, order)$}
    \State $g.append(c*)$
    \State $P \gets empty$
    \State $P(R).append(p(R)[c*])$
     \State $P(B).append(p(B)[c*])$
      \State $P(Y).append(p(Y)[c*])$
       \State $P(A).append(p(A)[c*])$
   \State $k \gets k-1$
   \State $B \gets w.copy$
   \State $w'\gets w$
   \While{$k \geq 0$}
        \State $order.remove(c*)$
        \State $B' \gets empty$
        \For{$b \in B$}
            \If{$b(c*) = R$}
                \State $B'.add(b)$
            \EndIf
        \EndFor
        \State $B \gets B'$
        \State{$p \gets \mathtt{Card Probabilities}(B,order.copy,k)$}
        \State{$v* \gets 0$}
        \State{$c* \gets null$}
        \For{$c \in order$}
            \If{$p(r)[c] > u$}
                \State $v \gets p(r)[c]*v_r + p(b)[c)*v_b + p(y)[c]*v_y + p(a)[c]*v_a$
                 \If{$v > v*$}
                    \State $v* \gets v$
                    \State $c* \gets c$
                 \EndIf{}
            \EndIf
        \EndFor
        \If{$c* = null$}
            \State $k \gets 0$
        \Else
            \State $g.append(c*)$
               \State $P(R).append(p(R)[c*])$
     \State $P(B).append(p(B)[c*])$
      \State $P(Y).append(p(Y)[c*])$
       \State $P(A).append(p(A)[c*])$
        \EndIf{}
        \State $k \gets k-1$
   \EndWhile
   \State{\Return $g*$}
\end{algorithmic}
\end{algorithm}

\begin{algorithm}[H] 
\caption{First Guess}
\label{alg:first-guess}
\begin{algorithmic}
    \Require Card probabilities $p$, ordered cards $order$
    \State{$v* \gets -\inf$}
    \State{$c* \gets null$}
    \For{$c \in order$}
        \If{$p(r)[c] > u$}
            \State $v \gets p(r)[c]*v_r + p(b)[c)*v_b + p(y)[c]*v_y + p(a)[c]*v_a$
             \If{$v > v*$}
                \State $v* \gets v$
                \State $c* \gets c$
             \EndIf{}
        \EndIf
    \EndFor
    \If{$c* = null$}
         \For{$c \in order$}
        
                \State $v \gets p(r)[c]*v_r + p(b)[c)*v_b + p(y)[c]*v_y + p(a)[c]*v_a$
                 \If{$v > v*$}
                    \State $v* \gets v$
                    \State $c* \gets c$
                 \EndIf{}
        \EndFor
    \EndIf
    \State{\Return $c*$}
\end{algorithmic}
\end{algorithm}

\begin{algorithm}[H] 
\caption{Card Probabilities}
\label{alg:card-prob}
\begin{algorithmic}
    \Require Sample of world states $W$, distribution over world states $p(w)$,
    clue $l_t$, unknown cards on board $B$, skip threshold $s$, ordering of cards $order$, number of closest cards $k$ 
    \State{$sum \gets 0$}
    \For{$w \in W$}
        \State{$sum \gets sum + p(w)$}
    \EndFor
    \For{$w \in W$}
        \State{$p(W) \gets p(w)/sum$}
    \EndFor
    \For{$c \in B$}
        \State{$p(r)[c] \gets 0$}
        \State{$p(b)[c] \gets 0$}
        \State{$p(y)[c] \gets 0$}
        \State{$p(a)[c] \gets 0$}
        \For{$w \in W$}
            \State{$p(w(c))[c] \gets p(w(c))[c] + p(w)$}
        \EndFor
        \If{$p(r)[c] \leq s$}
            \State{$order.remove(c)$}
        \EndIf{}
    \EndFor{}
    \For{$i < k$}
        \State{$c \gets order[i]$}
        \State{$p(r)[c] = 1$}
        \State{$p(b)[c] = 0$}
        \State{$p(y)[c] = 0$}
        \State{$p(a)[c] = 0$}
    \EndFor
    \State{\Return $p$}
\end{algorithmic}
\end{algorithm}

\begin{algorithm}[H] 
\caption{Update Bayesian Guesser}
\label{alg:update-bg}
\begin{algorithmic}
    \Require Set of spymasters $M$, prior probability distribution over spymasters $q(M)$, posterior distribution of guesses $p(M)$, card index $i$, probability of outcomes for each card $P(o)[i]$, observed card team $o$
    \If{$P(o)[i] = 0$}
        \For{$m \in  M$}
            \State $p(m) \gets q(m)$
        \EndFor
    \EndIf{}
\end{algorithmic}
\end{algorithm}

\section{The Bayesian Spymaster} \label{app:spymaster}
In this section we give the complete algorithm for the Bayesian Spymaster, which has the following variables and parameters:

\begin{itemize}
    \item Set of possible guessers $\hat{G}$%\footnote{note - as previously defined, the guesser function is technically a map from sequence of clues to actions, but we will treat as if it maps clues to actions by having the previous clues in sequence be recorded using the update models operation, changing the value of g} 

\item Prior Distribution of Guessers: $\hat{P}(g)$

\item Posterior Distribution of Guessers at turn $t$: $\hat{P_t}(g)$

\item Assumed noise $\hat{\eta}$  

\item Number of samples taken: $s$ 

\item Value of observed guess: $v(\gamma^*, A)$

\item Likelihood estimation of observed guess at turn $t$: $\hat{P}_t(\gamma^*\vert g)$

\item Pseudo-counts of $\gamma^*$ given $g$, $l$, and $n$ at turn $t$: $h_t(g,l, n)$[$\gamma^*$]

\end{itemize}

Algorithms \ref{alg:bsm-first-turn} and \ref{alg:bsm-subs-turn} should be called on the first turn and other turns, respectively, to update the beliefs of the Bayesian spymaster. 
Algorithm \ref{alg:get-clue} is called on each turn to get the clue from the spymaster, and Algorithms \ref{alg:get-sum-distance} and \ref{alg:obs-act} are sub-algorithms that are used in that process.

\begin{algorithm}[H]
\caption{First Turn, initialize For turn $t = 0$}
\label{alg:bsm-first-turn}
\begin{algorithmic}
    
    \State $\hat{P}_0 \gets \hat{P}$
    \State \Return $\mathtt{getClue}()$

\end{algorithmic}
\end{algorithm}

\begin{algorithm}[H]
\caption{Subsequent Turns, update beliefs for each subsequent turn $t > 0$ }
\label{alg:bsm-subs-turn}
\begin{algorithmic}
    \Require observed guess $\gamma*$
    \State $\hat{P}_{t}(\gamma^*) \gets \hat{P}_{t-1}(\gamma^*) \cdot \hat{P}_{t-1}(\gamma^* \vert g)$ %no l because it's been absorbed into the updated model, marginalization term ignored because only relative value matters when making decisions and it's never sampled from, only used as a weight
    \State \Return $\mathtt{getClue}()$

\end{algorithmic}
\end{algorithm}

\begin{algorithm}[H]
\caption{Get Clue}
\label{alg:get-clue}
\begin{algorithmic}
    \Require Possible clues $L$, count of unguessed red cards $\rho$
    \State $l^+ \gets null$
    \Comment{Best Clue}
    \State $E(l^+) \gets 0$
    \Comment{Expected Value}
    \State $Q^+ \gets \inf$
    \For {each clue $l \in L$}
        \For {each $n \leq \rho$} 
            \State $E(l,n) \gets 0$
            \State $Q \gets \mathtt{GetSumDistance}(l,n, g)$
            \For {each guesser $g \in \hat{G}$} 
                \State $h_t(g,l, n) \gets 1$
                \For {$s$ samples}
                    \State $\hat{z} \sim \mathcal{N}(0,\hat{\eta})$
                    \State $l^* = l + \hat{z}$
                    \State $\gamma \gets g(l^*, n)$
                    \State $\gamma^* \gets \mathtt{GetObservedAction}(\gamma)$
                    \State $E(l,n) \gets E(l,n) + v(\gamma, A) \cdot \hat{p}_t(g)$
                    \State $h_t(g,l,n)[\gamma^*] \gets h(g,l,n)[\gamma^*] + 1$
                \EndFor
            \EndFor
        \EndFor
        \If {$E(l,n) > E(l^+)$}
            \State $l^+ \gets (l,n)$
            \State $E(l^+) \gets E(l,n)$
            \State $Q^+ \gets Q$
        \ElsIf{$E(l,n) = E(l^*)$} %edge case to ensure it generalizes previous model
            \If {$Q < Q^+$}
                \State $l^+ \gets (l,n)$
                \State $E(l^+) \gets E(l,n)$
                \State $Q^+ \gets Q$
            \EndIf{}
        \EndIf{}
    \EndFor{} 
    \State Update $G$ with $l^+$
    \State \Return $l^+$
\end{algorithmic}
\end{algorithm}

\begin{algorithm}[H]
\caption{Get Observed Action}
\label{alg:obs-act}
\begin{algorithmic}
    \Function{$\mathtt{getObservedAction}$}{}
    \State $\gamma^* \gets []$
    \For{ $c$ in $\gamma$}
    \Comment{Iterate in order}
        \State $\gamma^*.\mathtt{insert}(c)$
        \If {$c \neq r$}
            \State \Return $\gamma^*$
        \EndIf{}
    \EndFor{} 
    \State \Return $\gamma^*$
    \EndFunction
\end{algorithmic}
\end{algorithm}

\begin{algorithm}[H] 
\caption{Get Sum Distance}
\label{alg:get-sum-distance}
\begin{algorithmic}
    \Require What is the input to this function? 
    \Function{$\mathtt{GetSumDistance}$}{}
    \State $Q \gets 0$
    \For{ $c$ in $\gamma$}
    \Comment{Iterate in order}
        \If {$c \neq r$}
            \State \Return $Q$
        \EndIf{}
         \State $Q \gets Q + ||c - l||_2$
    \EndFor{} \\
    \State \Return $Q$
    \EndFunction

\end{algorithmic}
\end{algorithm}

\section{Additional Details} 
This section provides some additional details on a few aspects of the Bayesian agent framework.

\subsection{Utility Heuristic}

The following utility heuristic is used: 

$v(\gamma,A) = \sum{v(A(\gamma_i))} - 1$ 
% minus one is for end of turn, ensures other cards are weighed properly for solitaire

$v(r) = 1$,
$v(b) = -1$, %note in previous paper the other cards are blue and thus avoided 
$v(y) = 0 $, %guessing them in and of itself is harmless in solitaire, but it ends the turn
$v(a) = -|R|$

%The final index on the sum is the index of the first non-red card in the list, the length of the list, or the number of cards remaining, whichever is defined and smallest. 

The rational behind this heuristic is that it calculates the marginal contribution to the score at the end of the game from current term assuming a particular variant of the solitaire Codenames. In this version, the game is played until all red cards are guessed, and then the score is given as the total number of red cards minus the number of blue cards guessed, minus the total number of red cards if the assassin was guessed, and minus the number of turns. 

In cases where the game is won, this gives the same score as the solitaire variant included in the official rules of Codenames. The difference between this variant and the official rules is that in the official rules, the codemaster chooses a blue card to remove each turn, while in this variant the blue cards remain unless guessed and the impact of turns taken is only calculated at the end of the game. Any positive score corresponds with a won game, while negatives and zero are lost games. 

If there is no noise and only one guesser, this heuristic will result in the Bayesian codemaster giving the same clue as the corresponding level-$k$ guesser because it will always return $|l|-1$ since no non-red cards will be expected to be guessed, and $|l|-1$ has the same ordinal relationships as $|l|$. 

\subsection{Estimating Likelihood}

To estimate likelihood, we assume a symmetric Dirichlet distribution with a pseudo-count of one for all guesses as the prior for the likelihood. 
Note the guesses here are the guesses observed by the codemaster, not the guesses intended by the guesser, as the guesser will not be able to execute their intended guess if they guess a card that is not on their team. The estimate used will be the mean of the posterior distribution. 
To calculate this, a map is defined from observed guesses to pseudo-counts. 
When ever a guess is observed during sampling, the pseudo-count is increased by one, with the value in the map being initialized to two if the guess was not already defined as a key value. When updating the codemaster's beliefs, the pseudo-count is used as the likelihood since it does not need to be normalized, with one being used when the key value was not found. We can summarize the implementation with the following formula:

$$
P_t(\gamma*|g) = h_t(g,l,n)
$$

\subsection{Optimization}
Without uncertainty, if there is a clue $(l,n)$ so that $(l,n)$ results in guessing only cards guessed on the team but $(l,n+1)$ results in guessing a card not on the team, then $(l,n)$ will have a greater or equal expected value to all $(l,m)$ where $m > n$. This allows search to be optimized by only stopping testing increasing numbers for a clue once n has been found. However, with uncertainty such an n is no longer defined. In particular, empirically finding an n where a card not on the team was guessed does not imply the clue with n+1 will result in a card not on the team being guessed. This is because either the guesser may change, or the clue will change when noise is added. With this in mind we propose the following filter for optimization. Test $(l,n)$ without adding noise and see what each guesser would guess. If all guessers guess a card not on the team, stop increasing n and move on to the next clue vector. This optimization is preferred first because we need to ensure a clue is returned, and it's possible that even if there is individually a clue closest for each card on the players' team for each guesser, that there is none that is closest simultaneously for all guessers, which would result in all clues being skipped if the clue had to return a guess with no cards on the other team for all guessers. Second, if the clue without noise leads to cards not on the team being guessed for some guessers but not others, then increasing the number may increase the number of cards guessed on the team for guessers were only cards on the team were guessed without leading to any change for the other guessers, increasing the expected utility. Finally, while it is possible that adding noise can move a clue away from the card not on the team that would have been guessed in order to increase the number of cards on the team guessed, this outcome is unlikely since adding noise is as likely to perturb the clue in the direction of that card as to perturb it in the opposite direction. We found that in practice including this optimization results in a much shorter running time without impacting results, so we have included it in the experiments we run. Algorithm \ref{alg:opt-get-clue} depicts all of the details. 

\begin{algorithm}
\caption{Optimized Get Clue}
\label{alg:opt-get-clue}
\begin{algorithmic}
    \Require Possible clues $L$, count of unguessed red cards $\rho$
    \State $l^+ \gets null$
    \Comment{Best Clue}
    \State $E(l^+) \gets 0$
    \Comment{Expected Value}
    \State $Q^+ \gets \inf$
    \For {each clue $l \in L$}
        \State $m \gets \rho$
        \For {each $n \leq m$}
           \State $viable \gets false$
            \For {each guesser $g \in \hat{G}$}
                \State$\gamma \gets g(l,n)$
                \State $\gamma^* \gets 
                \mathtt{GetObservedAction}(\gamma)$
                \If{$\gamma^* = \gamma$}
                    \State $viable \gets true$
                \EndIf{}
            \EndFor
             \If{Not $viable$}
                \State $m \gets n - 1$
              \EndIf{} 
            \If{$n \leq m$}  
                \State $E(l,n) \gets 0$
                \State $Q \gets \mathtt{GetSumDistance}(l,n, g)$
                \For {each guesser $g \in \hat{G}$} 
                    \State $h_t(g,l, n) \gets 1$
                    \For {$s$ samples}
                        \State $\hat{z} \sim \mathcal{N}(0,\hat{\eta})$
                        \State $l^* = l + \hat{z}$
                        \State $\gamma \gets g(l^*, n)$
                        \State $\gamma^* \gets \mathtt{GetObservedAction}(\gamma)$
                        \State $E(l,n) \gets E(l,n) + v(\gamma, A) \cdot \hat{p}_t(g)$
                        \State $h_t(g,l,n)[\gamma^*] \gets h(g,l,n)[\gamma^*] + 1$
                    \EndFor
                \EndFor
            \EndIf{}
        \EndFor
        \If {$E(l,n) > E(l^+)$}
            \State $l^+ \gets (l,n)$
            \State $E(l^+) \gets E(l,n)$
            \State $Q^+ \gets Q$
        \ElsIf{$E(l,n) = E(l^*)$} %edge case to ensure it generalizes previous model
            \If {$Q < Q^+$}
                \State $l^+ \gets (l,n)$
                \State $E(l^+) \gets E(l,n)$
                \State $Q^+ \gets Q$
            \EndIf{}
        \EndIf{}
    \EndFor{} 
    \State Update $G$ with $l^+$
    \State \Return $l^+$
\end{algorithmic}
\end{algorithm}

Source code can be found here: \url{https://github.com/Ganondox/bayesian-codenames}

\end{appendices}

\end{document}